\documentclass[letterpaper,journal]{IEEEtran}
\usepackage{threeparttable}
\usepackage[T1]{fontenc}
\usepackage[utf8]{inputenc}
\usepackage{cite}
\usepackage{amsmath,amssymb,amsfonts}
\usepackage{mathrsfs}
\usepackage{textcomp}
\usepackage{graphicx}
\usepackage[table]{xcolor}
\usepackage{array,booktabs,tabularx,multirow,makecell}
\usepackage{comment}
\usepackage{stfloats}
\usepackage{url}
\usepackage[hidelinks]{hyperref}
\definecolor{softblue}{RGB}{235,240,255}
\definecolor{meddarkblue}{RGB}{31,78,121}
\definecolor{medlightblue}{RGB}{242,246,250}
\definecolor{newcolor}{rgb}{.8,.349,.1}
\begin{document}
\bstctlcite{TMIreferencecontrol}
\title{Decoupling Disease, Covariates, and Individual Variability: A Unified Disentanglement Framework for Medical Image Classification}

\author{Shengjie Zhang, Jinglin Zhang, Zhuangzhuang Jiang, Ziqi Yu, Yipin Zhang,\\
Qi Zhang, Xiang Chen, Haibo Yang, Fei Gao, Longbiao Cui, Yuan Zhou, Xiao-Yong Zhang,\\
and Alzheimer's Disease Neuroimaging Initiative%
\thanks{Shengjie Zhang, Jinglin Zhang, and Zhuangzhuang Jiang contributed equally to this work. (Corresponding authors: Yuan Zhou and Xiao-Yong Zhang)}%
\thanks{S. Zhang, Z. Yu, Q. Zhang, and X.-Y. Zhang are with the Department of Radiology \& Faculty of Medical Imaging Technology, Ruijin Hospital, College of Health Science and Technology, Shanghai Jiao Tong University School of Medicine, Shanghai 200025, China. S. Zhang, Z. Yu, and X.-Y. Zhang are also with the Clinical Neuroscience Center, Ruijin Hospital, Shanghai Jiao Tong University School of Medicine, Shanghai 200025, China, and the Shanghai Key Laboratory of Child Brain and Development, Shanghai Children's Medical Center, Shanghai Jiao Tong University School of Medicine, Shanghai 200127, China (e-mail: \nolinkurl{zhangxiaoyong@sjtu.edu.cn}).}%
\thanks{J. Zhang, Z. Jiang, and Y. Zhou are with the School of Data Science, Fudan University, Shanghai 200433, China (e-mail: \nolinkurl{yuanzhou@fudan.edu.cn}).}%
\thanks{Y. Zhang, X. Chen, and H. Yang are with the Institute of Science and Technology for Brain-Inspired Intelligence, Fudan University, Shanghai 200433, China.}%
\thanks{F. Gao is with the Department of Radiology, Shandong Provincial Hospital Affiliated to Shandong First Medical University, Jinan 250031, China.}%
\thanks{L. Cui is with the Department of Clinical Psychology, School of Medical Psychology, Air Force Medical University, Xi'an 710032, Shaanxi, China.}%
}

\maketitle

\begin{abstract}
Accurately isolating disease-related features from confounding covariates (e.g., age, gender, site) and individual variations remains a fundamental challenge in medical image classification. Traditional regression-based approaches may ignore non-linear relations between image features and true covariates. To overcome this issue, we present a generalized Medical Imaging Disentanglement Learning (MedIDL) framework. MedIDL maps image features into three mutually orthogonal latent spaces through specialized disentanglement heads: a disease classification head guided by a supervised loss, a covariate-alignment head constrained by cross-subject similarity matching, and a Gaussian head absorbing individual variations. We evaluated our framework across 7 datasets encompassing diverse imaging modalities. MedIDL outperforms state-of-the-art supervised and self-supervised classification methods in accuracy across all datasets. Association analyses demonstrate that MedIDL successfully isolates target-specific latent representations. Gradient-based interpretability mappings localize pathognomonic patterns aligning with established clinical literature.
\end{abstract}

\begin{IEEEkeywords}
Disentanglement Learning, Medical Image Classification, Feature Distillation, Covariate Confounding.
\end{IEEEkeywords}

\section{Introduction}
\IEEEPARstart{D}{eep} learning has revolutionized medical image analysis by learning representations directly from data \cite{thakur2024deep}. Medical imaging spans a wide spectrum of modalities, from 2D radiographs and 3D volumetric magnetic resonance imaging (MRI) to brain networks derived from resting-state functional MRI (rs-fMRI) or diffusion tensor imaging (DTI). These diverse data structures provide complementary, multi-scale views of human physiology and pathology \cite{alharbi2025deep}. For example, rs-fMRI captures dynamic blood-oxygen-level-dependent (BOLD) fluctuations that reflect functional connectivity, while DTI and structural MRI (sMRI) reveal the underlying white-matter connectivity and gray-matter morphology respectively \cite{mohammadi2024graph}. Accurately classifying patients from healthy controls (HCs) using these diverse imaging sources can diagnose patients at an early stage and reveal disease-related changes \cite{han2025incomplete}.

\begin{figure}[!t]
    \centering
    \includegraphics[width=1.0\linewidth]{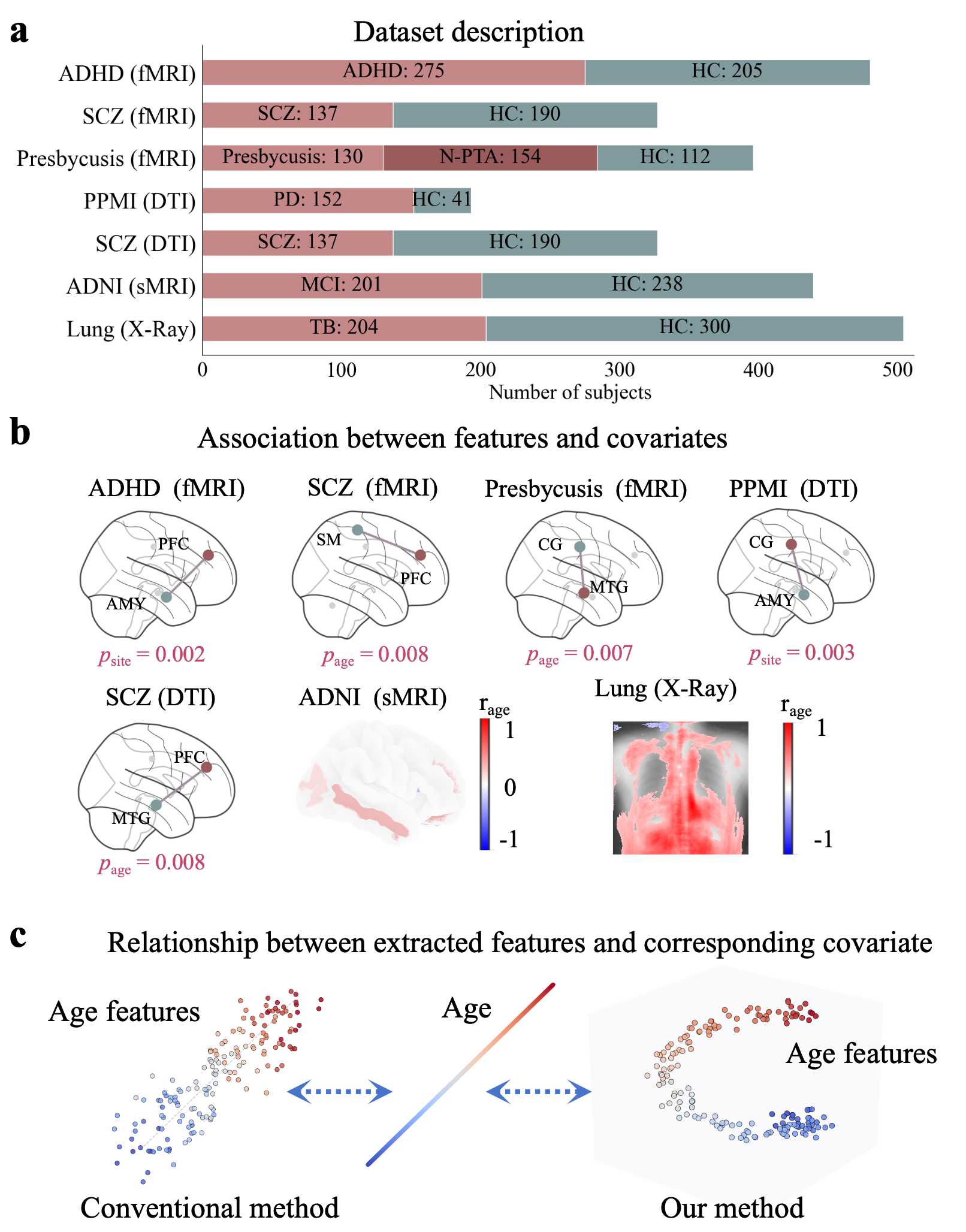}
    \caption{Overview of the datasets, covariate association, and advantage of the proposed framework. (a) illustrates the dataset description, showing the sample sizes across diverse modalities. (b) visualizes the most significantly associated biomarker(s) for a covariate (e.g. age or scanner site) within the HCs of a dataset. The image biomarkers are functional/structural connectivity for fMRI/DTI, intensity for sMRI and X-ray. The $p$-values were false discovery rate (FDR)-corrected in each dataset. (c) conceptually illustrates the latent representation spaces, highlighting that conventional methods typically enforce linear relationships between the extracted covariate features and true covariates, whereas our framework allows for non-linear manifold alignment.}
    \label{fig:motivation}
\end{figure}

In this task, a fundamental challenge exists: deep learning models struggle to disentangle disease-specific pathological signals from confounding covariates (such as age, gender, acquisition site, education level, and scanner type) and individual variability \cite{kheiri2025deceptive}. In some real-world preprocessed datasets (Fig.~\ref{fig:motivation}a), image features are highly correlated with various covariates. For example, age is positively associated with the prefrontal cortex-amygdala connectivity in fMRI of young healthy individuals in the Attention Deficit Hyperactive Disorder (ADHD)-200 dataset and positively associated with the temporal lobe intensity in sMRI (T1) of healthy individuals in the Alzheimer's Disease Neuroimaging Initiative (ADNI) (Fig. \ref{fig:motivation}b). Besides these covariate associations, individual variability also exists. For example, HC subjects who share identical demographic profiles exhibit substantial inter-individual differences in imaging appearance \cite{hendriks2026impact}. As a result, standard convolutional neural networks (CNNs), vision transformers (ViTs), and graph neural networks (GNNs) could inadvertently encode and memorize these nuisance variables. This entanglement leads to biased representations, poor cross-site generalizability, and inflated performance on in-distribution data \cite{boland2024there}.

Existing strategies for handling this entanglement remain insufficient. Traditional methods typically linearly regress image features on covariates to adjust the features to the center of covariates, assuming that these exists a linear relationship~\cite{alfaro2021confound}. Normative modeling extends the linear relationship by using Gaussian process regression to predict image features from covariates~\cite{marquand2016understanding}. Residuals from the prediction are considered as features deprived of covariate effects. With the emergence of deep learning, adversarial learning and domain-adversarial neural networks have been employed to suppress demographic biases \cite{zhao2025dfca} while regression has also been attempted to isolate covariates such as age \cite{yan2024dual}. These approaches typically address only a subset of confounders or operate under restrictive linear assumptions (Fig.~\ref{fig:motivation}c), 
which may overly constrain the representation space and fail to capture complex, non-linear interactions between features and true covariates.

Furthermore, existing methods are hindered by a pervasive architectural compartmentalization. Specifically, GNNs operate almost exclusively on non-Euclidean topologies for functional and structural brain connectomes \cite{mohammadi2024graph}, while CNNs and ViTs are confined to the Euclidean grid structures characteristic of 2D planar or 3D volumetric tensors. This pre-empts the formulation of a unified, geometry-agnostic representation paradigm capable of assimilating heterogeneous data topologies within a single optimization landscape. Consequently, adapting to different imaging modalities necessitates modality-specific network re-engineering and isolated optimization pipelines.

To overcome these limitations, we propose a generalized Medical Imaging Disentanglement Learning (MedIDL) framework that explicitly decomposes the latent representation of \emph{any} medical imaging input into three complementary components: (i) disease-related features that separate HCs from patients, (ii) covariate-related features that capture demographic and acquisition-related effects, and (iii) individual variations that account for residual subject-specific differences. Inspired by contrastive variational autoencoders \cite{aglinskas2022contrastive}, MedIDL incorporates a novel covariate alignment scheme to isolate covariate-related features without assuming a linear relationship with the true covariate. Specifically, covariate-related features are learned by explicitly aligning a similarity matrix between latent embeddings of subjects with the ground-truth covariate similarity matrix, enabling the covariate features to preserve neighborhood proximity while being flexible enough to form any 1-dimensional manifold in the feature space for a covariate (Fig.~\ref{fig:motivation}c). Furthermore, disease-related features are isolated through supervised classification, while an independent individual head captures the remaining variability. 

To validate the proposed framework, we conduct comprehensive evaluations across seven medical datasets, covering functional networks (fMRI), structural networks (DTI), 3D anatomical volumes (sMRI), and 2D radiographs (X-ray) (Fig. \ref{fig:motivation}a). The main contributions of this work are threefold:
\begin{itemize}
    \item A unified disentanglement framework: We present the first end-to-end model capable of decomposing latent features into disease-related, covariate-related, and individual-variation components without requiring restrictive linear assumptions.
    \item Modality-adaptable architecture: We design a lightweight encoder-adaptive mechanism that enables adaptation across 2D planar images, 3D volumetric scans, and graph-structured brain networks.
    \item Extensive multimodal validation: We conduct comprehensive experiments on 7 datasets spanning diverse modalities and diseases including ADHD-200, Schizophrenia (SCZ), Presbycusis, Parkinson's Progression Marker Initiative (PPMI), ADNI, and Lung.
\end{itemize}
Preliminary work was published in 2025 International conference on Medical Image Computing and Computer-Assisted Intervention (MICCAI)~\cite{zhang2025graph}. 
\begin{figure*}[!t]
     \centering
    \includegraphics[width=\textwidth]{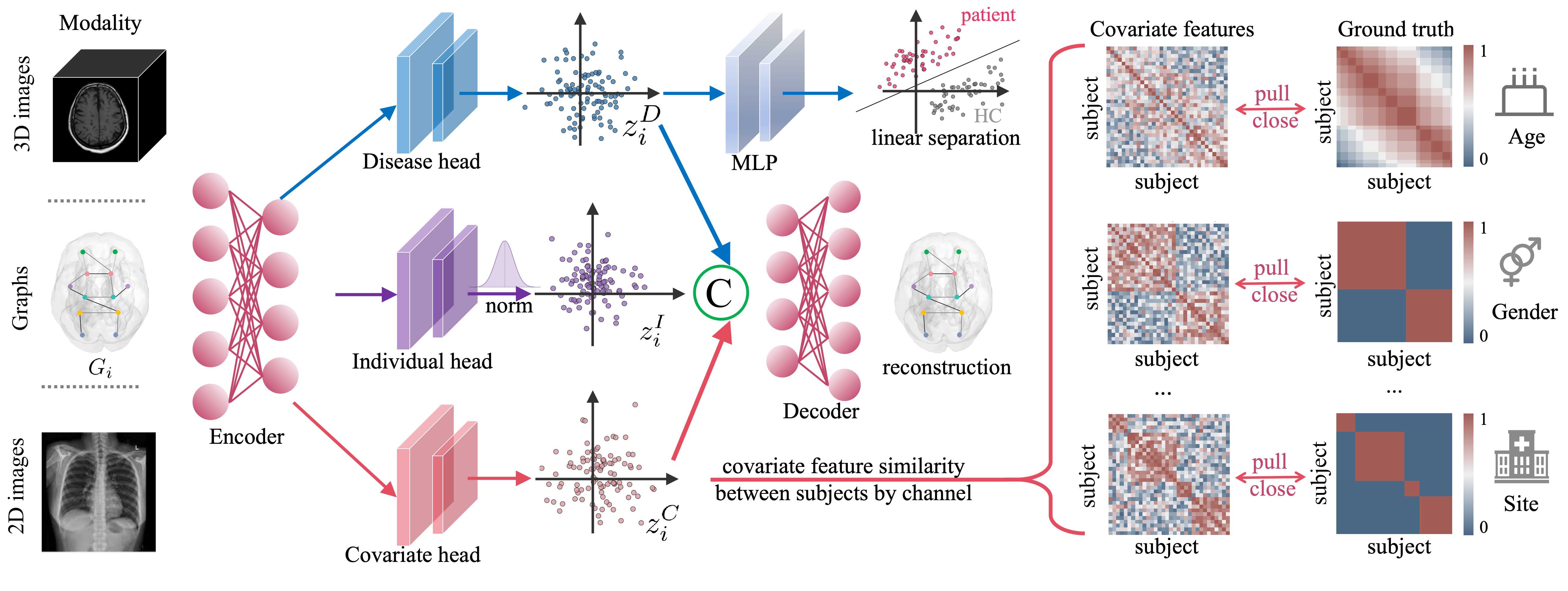}
\caption{Overall architecture of the proposed disentangled representation learning framework. For any given input modality (e.g.,  3D images, brain graphs, or 2D images), the data is processed by an encoder and subsequently projected into three separate latent spaces via dedicated heads. The disease head extracts disease-specific features ($z_i^D$) optimized for separating patients from HCs. The individual head captures normalized, individual-specific latent variations ($z_i^I$). The covariate head extracts demographic and site-related features ($z_i^C$), which are supervised by aligning the channel-wise subject similarity matrices with the corresponding ground truth similarity matrices from the covariates (e.g., age, gender, and site). Finally, the three disentangled components are concatenated and fed into a decoder to reconstruct the original input.}
\label{fig:architecture}
\end{figure*}

\section{Related work}
\label{sec:related_work}

\subsection{Medical image classification algorithms}
Over the past decade, computer-aided diagnosis has transitioned from localized spatial feature extraction to global, multi-modal, and topology-aware classification frameworks. Early milestones pioneered by Li et al. \cite{li2014medical} deployed shallow CNNs to replace hand-crafted descriptors, a paradigm subsequently extended by Zhang et al. \cite{zhang2019medical}. As reviewed by Chen et al. \cite{chen2025review}, while CNNs remain effective for localized lesion extraction, their stationary receptive fields inherently restrict macroscale long-range dependency modeling. To bridge this contextual gap, hybrid ViTs emerged. Manzari et al. \cite{manzari2023medvit} developed MedViT to combine convolutional efficiency with global self-attention, while Wu et al. \cite{wu2023ctranscnn} engineered CTransCNN to seamlessly integrate parallel CNN and Transformer pathways for multi-label diagnosis. Further optimizing self-attention, Chowdary and Yin \cite{chowdary2024med} introduced Med-former with specialized tokenization to capture complex multi-sequence patterns. Recent frameworks migrated to frequency domains and non-Euclidean topological spaces. For example, Zhang et al. \cite{zhang20223d} proposed a 3D global Fourier network, utilizing 3D fast Fourier transforms to extract global descriptors from sMRI, while Li et al. \cite{li2021braingnn} introduced BrainGNN to leverage graph convolutions and pooling for functional connectivity classification. Despite the transition toward holistic and topology-aware architectures, these paradigms still struggle with inherent feature entanglement.

\subsection{Covariate removal strategy}

Traditionally, linear regression for removing the covariate effects has existed for a long time. For example, voxel-based morphometry (VBM) utilizes Generalized Linear Models (GLMs) to account for nuisance covariates \cite{ashburner2000voxel}, a practice also performed at the population scale by Smith et al. \cite{smith2015positive}. Explicit covariate elimination is widely formulated via regression-based residualization, where GLMs are fitted on HC subjects to yield orthogonalized residual features \cite{dukart2011age}. In addition, such a covariate-removal workflow is often paired with z-score normalization to remove the scale of different dimensions \cite{falahati2016effect}. Furthermore, advanced harmonization pipelines like ComBat \cite{fortin2018harmonization} extended standard GLMs via empirical Bayes estimation. Although highly efficient and interpretable, these traditional paradigms operate under rigid parametric assumptions and simplified linear constraints. Consequently, they may fail to model non-linear interactions between covariates and image features.

\subsection{Deep disentanglement learning in medical imaging}

Modern deep learning architectures have shifted toward disentangled representation learning (DRL) to explicitly model and isolate heterogeneous covariates within latent spaces. In this paradigm, treating nuisance variations as explicit conditioning factors or adversarial targets serves as an architectural form of feature factorization. For instance, Zhao et al. \cite{zhao2020training} introduced an adversarial learning framework by formulating a competitive game between a primary disease predictor and a confounder classifier. By training the predictor to actively deceive the classifier, the network learns imaging representations independent of site or demographic variations. Chartsias et al. \cite{chartsias2019disentangled} factorized medical images into mutually exclusive sub-spaces separating invariant anatomy from style factors, while Moyer et al. \cite{moyer2020scanner} advanced scanner-invariant representation learning for diffusion MRI by deploying information-theoretic boundaries to filter out site-specific variations from brain connectivity networks. 

Concurrently, deep contrastive learning and non-linear metric constraints have transformed how networks isolate target pathological variations from complex backgrounds. For example, Aglinskas et al. \cite{aglinskas2022contrastive} introduced a contrastive machine learning framework that successfully disentangled shared healthy population backgrounds from neuroanatomical alterations in Autism Spectrum Disorder (ASD). To capture dynamic developmental trajectories, Yu et al. \cite{yu2022longitudinal} developed a conditional intensive triplet network that embeds time-progressive covariates directly into a deep metric loss. Recently, Ouyang et al. \cite{ouyang2022disentangling} weakly supervised longitudinal MRIs to segregate normal aging from disease severity, and Maeng et al. \cite{maeng2025idenbat} proposed IdenBAT to decouple age from age-invariant morphological identity traits. Collectively, these deep covariate-modeling methodologies establish a solid paradigm for separating disease-specific traits from confounders. 

\section{Method}
\label{sec:method}

\subsection{General architecture}

Let the input for the \(i\)-th subject be \(x_i\), which can be a graph, a 3D volumetric image, or a 2D planar image. The MedIDL framework explicitly decomposes the latent representation of any input medical image into three independent components: (i) disease-related feature \(z_i^D \in \mathbb{R}^d\), (ii) covariate-related feature \(z_i^C\in \mathbb{R}^d\) and (iii) individual variation \(z_i^I\in \mathbb{R}^d\).
The framework consists of three modular stages (Fig.~\ref{fig:architecture}):
\begin{enumerate}
    \item A modality-aware encoder \(Enc(\cdot)\) that extracts features based on the specific input data structure and projects them into a latent space \(z_i = Enc(x_i) \in \mathbb{R}^d\).
    \item Three parallel disentanglement heads, \textit{disease head} \(g_D\), \textit{covariate head} \(g_C\), and \textit{individual head} \(g_I\), that disentangle the unified representation into three components:
   \[
   z_i^D = g_D(z_i), \quad z_i^C = g_C(z_i), \quad z_i^I = g_I(z_i).
   \]
    \item A shared latent decoder \(Dec\) that reconstructs the original input from the concatenated disentangled features, ensuring information preservation.
\end{enumerate}

Next, we give details on the input data structure, the encoder, the disentanglement heads, the decoder, and the loss function.

\subsection{Input data structure and encoder}


    For brain graphs from fMRI/DTI, we have  
    \[
    x_i = G_i = (A_i, X_i), \quad A_i \in \mathbb{R}^{N \times N}, \quad X_i \in \mathbb{R}^{N \times F},
    \]
    where \(N = 116\) is the number of the nodes from the Automated Anatomical Labeling (AAL) 1 atlas, \(A_i\) is the adjacency matrix, and \(X_i\) contains node features. The adjacency matrix is Pearson's correlation coefficient between regional BOLD signals for fMRI or mean streamline count for DTI. The node features are amplitude of low-frequency fluctuation (ALFF) in Slow-5, Slow-4, and classical bands (\(F = 3\)) for fMRI. For DTI, \(X_i\) encapsulates the concatenated 6-dimensional microstructural tensor profile comprising fractional anisotropy (FA), mean diffusivity (MD), axial diffusivity (AD), radial diffusivity (RD), tensor mode (MO), and trace (TR) (\(F = 6\)). We use the graph isomorphism network (GIN) as the encoder for this type of data.

    For 3D volumetric MRI, we have  
    \[
    x_i = X_i \in \mathbb{R}^{H \times W \times D \times C},
    \]
    where \(H = W = D = 96\) and \(C = 1\), $x_i$ represents the processed T1-weighted MRI following a pipeline of skull-stripping, MNI-152 template alignment through affine registration, intensity normalization, center-cropping and scaling. We use the Swin-ViT as the encoder for this type of data.

    For 2D planar X-ray, we have  
    \[
    x_i = X_i \in \mathbb{R}^{H \times W \times C},
    \]
    where \(C = 1\), $x_i$ represents a grayscale chest radiograph, resized to \(224 \times 224\). We use a 2D residual CNN-based (Res-Conv) neural network as the encoder.

All encoders output a fixed \(d = 64\)-dimensional latent vector \(z_i \in \mathbb{R}^{d}\). This creates a modality-agnostic information bottleneck for the downstream disentanglement heads. Detailed architectural configurations for each encoder are provided in Appendix A.

\subsection{Disentanglement heads}

The disentanglement heads $g_D$, $g_C$, and $g_I$ are implemented as independent 2-layer multi-layer perceptrons (MLPs) with ReLU activations. Each head maps the representation $z_i$ into a $d$-dimensional latent vector, yielding $z_i^D, z_i^C, z_i^I \in \mathbb{R}^d$, respectively. 

\subsubsection{Covariate head}

The covariate head $g_C$ projects intermediate representations into a $d$-dimensional covariate feature space, yielding $z_i^C \in \mathbb{R}^d$. To align the latent space with the non-imaging covariates, we evaluate pairwise similarity matrices and force the feature-level similarity matrix to be close to the ground truth similarity matrix for each covariate.

Specifically, for a given covariate $k \in \mathcal{C}$ (e.g., $k \in \{\text{age}, \text{gender}, \text{site}\}$), we compute the feature-level similarity matrix $\hat{S}^{(k)} = [\hat{S}_{ij}^{(k)}]_{B \times B}$ over a mini-batch of size $B$ using a scale-invariant, lightweight linear projection $\phi_k(z_i^C)$:
\begin{equation}
\hat{S}^{(k)}_{ij} = \frac{\phi_k(z_i^C)^\top \phi_k(z_j^C)}{\|\phi_k(z_i^C)\|_2 \|\phi_k(z_j^C)\|_2}.
\end{equation}
The ground-truth covariate similarity matrix $S^{(k)} = [S^{(k)}_{ij}]_{B \times B}$ is defined based on the covariate type:
\begin{itemize}
    \item Continuous variables (e.g., age $a_i$):
To handle potential scaling issues, a continuous covariate $a_i$ is first linearly mapped into the range $[0, 1]$ over all the subjects in the training set, yielding $\tilde{a}_i$. The similarity is then defined as:
    \begin{equation}
    S^{(\text{age})}_{ij} = 1 - |\tilde{a}_i - \tilde{a}_j|.
    \end{equation}
    \item Categorical variables (e.g., gender or site $s_i$):
    \begin{equation}
    S^{(\text{site})}_{ij} = \mathbb{I}(s_i = s_j),
    \end{equation}
    where $\mathbb{I}(\cdot)$ is the indicator function.
\end{itemize}
The covariate alignment loss is then formulated as:
\begin{equation}
\mathcal{L}_{cov} = \frac{1}{\vert \mathcal{C} \vert}\sum_{k \in \mathcal{C}} \frac{1}{B^2} \left\| S^{(k)} - \hat{S}^{(k)} \right\|_F^2,
\end{equation}
where $\|\cdot\|_F$ denotes the Frobenius norm.

\subsubsection{Individual head}

To model individual variation without introducing complex optimization landscapes, we impose a parameter-free structural prior on the individual variations \(z_i^I\). Specifically, \(z_i^I\) is channel-wisely normalized (whitened) to approximate a standard Gaussian distribution in a batch:
\[
z_i^I \leftarrow \frac{z_i^I - \mu(z_i^I)}{\sigma(z_i^I) + \epsilon}, 
\]
where the mean operator $\mu$ and the standard deviation operator $\sigma$ operate in a batch, and \(\epsilon\) is a small constant for numerical stability. This simple yet effective whitening operation forces the network to absorb uninformative, random stochastic variations into \(z_i^I\).

\subsubsection{Disease head}

The disease-related features $z_i^D$ are fed into an MLP classifier:
\[
p_i=\operatorname{softmax}(h_D(z_i^D)).
\]
where $h_D$ is an MLP with 2 layers using the ReLU activation function and a hidden dimension of $d$. The supervised classification objective is:
\[
\mathcal{L}_{sup}
=
-\frac{1}{B}
\sum_{i=1}^{B}
\sum_{c=1}^{C}
y_{i,c}
\log(p_{i,c}),
\]
where $C$ is the number of classes, $y_i = [y_{i,1},\dots,y_{i,C}]^{\top}$ is the one-hot encoded class label. 

\subsection{Decoder and final loss function}

Finally, to guarantee minimal information loss during the disentanglement process, the concatenated features \([z_i^D; z_i^C; z_i^I] \in \mathbb{R}^{3d}\) are fed into a decoder to reconstruct the original image or node features \(\hat{X}_i = Dec([z_i^D; z_i^C; z_i^I])\), yielding the latent reconstruction loss:
\[
\mathcal{L}_{rec} = \frac{1}{B} \sum_{i=1}^{B} \|X_i - \hat{X}_i\|_2^2.
\]
The decoder $Dec$ is similarly a 2-layer MLP that maps the concatenated $3d$-dimensional vector back to the original input space. The hidden layer has a dimension of $d$ with a ReLU activation function. This reconstruction is efficient enough while ensuring that the disentangled features collectively preserve the full expressivity of the original input.

The total objective is collectively optimized end-to-end:
\[
\mathcal{L} = \mathcal{L}_{\sup} + \lambda_1 \mathcal{L}_{cov} + \lambda_2 \mathcal{L}_{rec},
\]
where \(\lambda_1\) and \(\lambda_2\) are fixed across all experiments. By subjecting each branch to mutually exclusive supervision signals (class labels, covariate similarities, whitening) while binding them together via the latent reconstruction loss, the framework effectively achieves robust implicit disentanglement.

\subsection{Implementation details}

The framework is implemented in PyTorch and trained on a single NVIDIA A6000 or V100 GPU.

The GIN encoder processes brain graphs using a 2-layer GIN with 64 hidden dimensions and ReLU activations, followed by global average pooling. 

The ViT encoder processes 3D MRI scans using $16 \times 16 \times 16$ patches and an 8-layer Transformer (6 heads, 768 hidden dimensions). Patch tokens are aggregated via global average pooling and linearly projected ($W_{\text{proj}} \in \mathbb{R}^{64 \times 768}$) to $z_i$.

The 2D CNN-based residual network extracts features from 2D X-ray radiographs across four residual down-sampling stages ($3 \times 3$ convolutions, batch normalization (BN), ReLU, and $1 \times 1$ projection shortcuts), followed by global average pooling.

We use the Adam optimizer (learning rate is \(1\times10^{-3}\) and weight decay is \(1\times10^{-5}\)), batch size $B=32$, and train for 300 epochs with early stopping on validation accuracy. The same hyperparameters, \(\lambda_1=1.0\), \(\lambda_2=0.6\), \(d=64\), \(\epsilon=10^{-4}\), are fixed across all datasets without per-dataset tuning.

\section{Results}
\label{sec:results}

\subsection{Datasets and preprocessing}

We used 7 datasets acquired from diverse multi-center clinical cohorts for evaluation, including 4 public datasets --- ADHD-200 (fMRI), PPMI (DTI), ADNI (sMRI), Lung (2D X-ray)~\cite{lung_data} --- and 3 private datasets --- SCZ (fMRI), SCZ (DTI), Presbycusis (fMRI). Both the functional and structural SCZ datasets are derived from the same multi-modal cohort, which was collected at Xijing Hospital, affiliated with the Fourth Military Medical University, China. They are treated as entirely independent benchmarks. The Presbycusis dataset was collected at Shandong Provincial Hospital Affiliated to Shandong First Medical University.

All datasets were uniformized into three canonical representation formats: non-Euclidean graphs (functional graph: ADHD-200, SCZ, Presbycusis; structural graph: PPMI, SCZ), 3D volumetric images (ADNI), and 2D planar images (Lung). All cohorts provide three clinical covariates for training --- age, gender, and site -- except for the Presbycusis and Lung datasets which contain only age and gender. Dataset details and preprocessing steps are left in Appendix~B.


\subsection{Classification accuracy}


Table~\ref{tab:comparison} summarizes the classification results on the fMRI (ADHD, SCZ, Presbycusis) and DTI (PPMI, SCZ) datasets under the evaluation strategy detailed in Appendix C. The proposed framework consistently achieves the highest accuracy and AUC across all five datasets. Notably, on the challenging SCZ fMRI and DTI classification tasks, our method yields accuracies of 70.02\% and 73.42\%, outperforming the strongest baseline (DMG) by absolute margins of 2.88\% and 3.33\%, respectively. Similarly, in the multi-class Presbycusis task, our model achieves an OvR-AUC of 87.05\%, demonstrating its capacity to handle multi-class scenarios. 


The evaluation on 3D/2D images is reported in Table~\ref{tab:comparison_2}. Again, our method establishes a new state-of-the-art on both 3D/2D datasets. On the ADNI dataset, it achieves an accuracy of 85.71\% and an AUC of 86.48\%, surpassing the highly competitive diffusion-based method DiffMed-v2 by 1.25\% in accuracy and 1.93\% in AUC. On the 2D X-ray lung-infection task, our framework reaches an accuracy of 87.60\%. 

\begin{table*}[!t]
\centering
\begin{threeparttable}
\caption{Classification performance on fMRI (ADHD-200, SCZ, Presbycusis) and DTI (PPMI, SCZ) connectomes.}
\label{tab:comparison}

\begin{tabular*}{\textwidth}{@{}c@{}}
\resizebox{\textwidth}{!}{%
\rowcolors{4}{softblue}{white}
\begin{tabular}{lcccccccccc}
\toprule
 & \multicolumn{6}{c}{fMRI} & \multicolumn{4}{c}{DTI} \\
\cmidrule(lr){2-7} \cmidrule(lr){8-11}
 & \multicolumn{2}{c}{ADHD}
 & \multicolumn{2}{c}{SCZ}
 & \multicolumn{2}{c}{Presbycusis}
 & \multicolumn{2}{c}{PPMI}
 & \multicolumn{2}{c}{SCZ} \\
\cmidrule(lr){2-3}
\cmidrule(lr){4-5}
\cmidrule(lr){6-7}
\cmidrule(lr){8-9}
\cmidrule(lr){10-11}
\multirow{-3}{*}{Method}
 & Accuracy \tnote{a}& AUC
 & Accuracy & AUC
 & Accuracy & OvR-AUC
 & Accuracy & AUC
 & Accuracy & AUC \\
\midrule
BrainGNN
 & 62.22 $\pm$ 3.87 & 63.04 $\pm$ 4.42
 & 64.86 $\pm$ 3.57 & 64.48 $\pm$ 2.73
 & 83.09 $\pm$ 4.41 & 84.02 $\pm$ 4.51
 & 67.54 $\pm$ 3.85 & 69.02 $\pm$ 4.17
 & 68.61 $\pm$ 3.26 & 69.22 $\pm$ 3.40 \\
NEGAT
 & 62.08 $\pm$ 2.94 & 62.95 $\pm$ 3.35
 & 63.33 $\pm$ 4.56 & 64.09 $\pm$ 3.91
 & 82.79 $\pm$ 3.85 & 82.95 $\pm$ 3.36
 & 69.31 $\pm$ 3.52 & 70.53 $\pm$ 3.92
 & 68.25 $\pm$ 3.82 & 68.72 $\pm$ 3.91 \\
GraphCL
 & 62.86 $\pm$ 4.71 & 62.21 $\pm$ 4.27
 & 65.31 $\pm$ 4.72 & 65.19 $\pm$ 4.40
 & 83.45 $\pm$ 4.49 & 82.50 $\pm$ 3.92
 & 70.65 $\pm$ 4.58 & 72.36 $\pm$ 4.20
 & 68.29 $\pm$ 3.41 & 68.88 $\pm$ 3.62 \\
JOAO
 & 61.92 $\pm$ 3.72 & 62.12 $\pm$ 3.84
 & 64.92 $\pm$ 5.11 & 64.80 $\pm$ 4.63
 & 83.15 $\pm$ 3.77 & 82.69 $\pm$ 4.43
 & 70.28 $\pm$ 3.59 & 71.92 $\pm$ 3.47
 & 68.37 $\pm$ 4.10 & 69.30 $\pm$ 3.86 \\
LaGraph
 & 63.88 $\pm$ 2.68 & 63.92 $\pm$ 2.63
 & 65.25 $\pm$ 3.83 & 64.46 $\pm$ 4.30
 & 83.85 $\pm$ 4.02 & 83.19 $\pm$ 3.46
 & 71.58 $\pm$ 3.72 & 72.20 $\pm$ 3.91
 & 69.24 $\pm$ 3.69 & 70.11 $\pm$ 4.03 \\
AGCL
 & 63.25 $\pm$ 4.18 & 62.63 $\pm$ 4.29
 & 67.25 $\pm$ 4.18 & 65.35 $\pm$ 3.64
 & 84.15 $\pm$ 4.17 & 83.55 $\pm$ 3.86
 & 72.48 $\pm$ 4.25 & 72.85 $\pm$ 3.74
 & 69.92 $\pm$ 3.88 & 70.16 $\pm$ 4.05 \\
GATE
 & 62.25 $\pm$ 3.84 & 62.71 $\pm$ 4.03
 & 66.28 $\pm$ 3.71 & 66.05 $\pm$ 4.12
 & 83.97 $\pm$ 3.47 & 83.27 $\pm$ 2.49
 & 72.18 $\pm$ 3.62 & 72.55 $\pm$ 4.10
 & 68.33 $\pm$ 3.92 & 69.81 $\pm$ 3.60 \\
BrainGCL
 & 62.05 $\pm$ 3.59 & 61.52 $\pm$ 3.70
 & 65.27 $\pm$ 2.97 & 64.31 $\pm$ 3.59
 & 82.85 $\pm$ 2.98 & 83.06 $\pm$ 4.14
 & 72.11 $\pm$ 4.13 & 72.46 $\pm$ 3.86
 & 68.25 $\pm$ 4.05 & 69.03 $\pm$ 3.72 \\
DMG
 & 62.57 $\pm$ 2.95 & 62.08 $\pm$ 3.51
 & 67.14 $\pm$ 3.95 & 68.52 $\pm$ 4.41
 & 84.08 $\pm$ 3.19 & 83.95 $\pm$ 4.39
 & 72.65 $\pm$ 3.50 & 73.07 $\pm$ 4.03
 & 70.09 $\pm$ 4.14 & 71.58 $\pm$ 3.95 \\
\midrule
Ours
 & \textbf{65.24} $\pm$ 3.17 \tnote{b}& \textbf{66.03} $\pm$ 4.05
 & \textbf{70.02} $\pm$ 4.06 & \textbf{71.05} $\pm$ 4.57
 & \textbf{86.22} $\pm$ 3.36 & \textbf{87.05} $\pm$ 5.04
 & \textbf{74.46} $\pm$ 3.52 & \textbf{74.91} $\pm$ 4.10
 & \textbf{73.42} $\pm$ 4.06 & \textbf{74.08} $\pm$ 4.02 \\
\bottomrule
\end{tabular}%
} \\
\end{tabular*}

\begin{tablenotes}[flushleft]
\footnotesize
\item[a] Test results are reported as mean $\pm$ standard deviation across 5 folds.
\item[b] The best result for each dataset and metric is in \textbf{boldface}.
\end{tablenotes}

\end{threeparttable}
\end{table*}

\begin{table}[!t]
\centering
\begin{threeparttable}
\caption{Classification performance on 2D/3D vision tasks.}
\label{tab:comparison_2}

\begin{tabular*}{\linewidth}{@{}c@{}}
\resizebox{\linewidth}{!}{%
\rowcolors{5}{white}{softblue}
\begin{tabular}{lcccc}
\toprule
 & \multicolumn{2}{c}{3D MRI} & \multicolumn{2}{c}{2D X-ray} \\
\cmidrule(lr){2-3} \cmidrule(lr){4-5}
 & \multicolumn{2}{c}{ADNI} & \multicolumn{2}{c}{Lung} \\
\cmidrule(lr){2-3} \cmidrule(lr){4-5}
\multirow{-3}{*}{Method} & Accuracy \tnote{a}& AUC & Accuracy & AUC \\
\midrule
v-CNN      & 80.53 $\pm$ 4.15 & 81.14 $\pm$ 3.58 & 82.35 $\pm$ 2.30 & 83.05 $\pm$ 1.96 \\
ResNet     & 81.32 $\pm$ 3.72 & 80.88 $\pm$ 4.05 & 82.95 $\pm$ 2.35 & 83.52 $\pm$ 2.06 \\
ViT        & 81.59 $\pm$ 3.90 & 82.22 $\pm$ 3.74 & 83.48 $\pm$ 2.02 & 84.15 $\pm$ 1.65 \\
GF-Net     & 83.07 $\pm$ 4.16 & 83.25 $\pm$ 3.10 & 85.05 $\pm$ 1.59 & 85.68 $\pm$ 2.15 \\
V-Mamba    & 82.29 $\pm$ 3.90 & 84.02 $\pm$ 4.07 & 85.28 $\pm$ 1.94 & 85.75 $\pm$ 2.23 \\
DiffMed-v2 & 84.46 $\pm$ 4.18 & 84.55 $\pm$ 3.29 & 86.26 $\pm$ 2.06 & 86.81 $\pm$ 1.45 \\
SimCLR     & 83.88 $\pm$ 2.95 & 84.09 $\pm$ 3.34 & 84.49 $\pm$ 1.45 & 84.50 $\pm$ 1.59 \\
ADIOS      & 83.58 $\pm$ 3.52 & 83.86 $\pm$ 3.75 & 85.15 $\pm$ 2.29 & 85.42 $\pm$ 1.88 \\
SGLA-Net   & 82.09 $\pm$ 3.85 & 83.11 $\pm$ 3.79 & 84.30 $\pm$ 1.92 & 84.85 $\pm$ 2.01 \\
\midrule
Ours & \textbf{85.71} $\pm$ 3.52 \tnote{b}& \textbf{86.48} $\pm$ 4.83 & \textbf{87.60} $\pm$ 3.42 & \textbf{87.91} $\pm$ 2.48 \\
\bottomrule
\end{tabular}%
} \\
\end{tabular*}

\begin{tablenotes}[flushleft]
\footnotesize
\item[a] Test results are reported as mean $\pm$ standard deviation across 5 folds.
\item[b] The best result for each dataset and metric is in \textbf{boldface}.
\end{tablenotes}

\end{threeparttable}
\end{table}

\subsection{Quantitative validation of feature disentanglement}

To verify that the proposed framework successfully isolates the target information in the latent space, we quantitatively evaluated the correlation between the learned latent representations and the ground truth covariates. Specifically, we computed the Kendall's Tau rank correlation coefficient between the five disentangled latent embeddings and the actual ground-truth covariates. For each test set, the similarity of a subject between all the other subjects is calculated for both the feature and the ground truth such that Kendall's Tau rank correlation coefficient can be calculated between these 2 similarity vectors, then the mean correlation coefficient is evaluated over all the subjects in this test set.

\begin{figure*}[!t]
    \centering
    \includegraphics[width=0.98\textwidth]{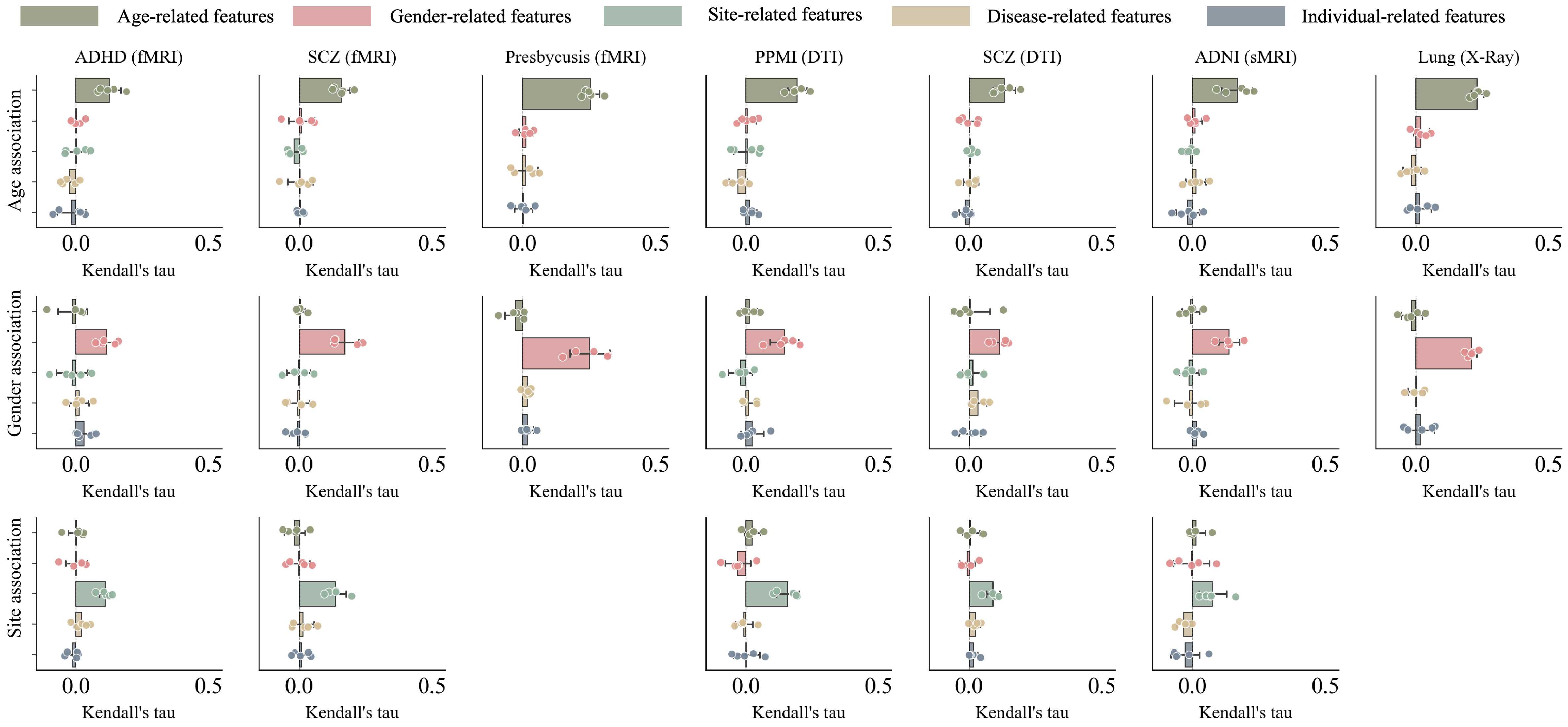}
    \caption{Quantitative validation of feature disentanglement using Kendall's Tau rank correlation across seven datasets. Each row represents a ground-truth covariate (age, gender, or site) for evaluating the correlation. A dot within a subplot denotes the mean correlation over all the subjects in a test set, where the correlation for a subject measures the order consistency between similarities from this subject to all the others measured by a distinct feature (by color) and similarities measured by the true covariate. 
    }
    \label{fig:kendall_tau}
\end{figure*}

The correlation analysis reveals a highly specific and orthogonal mapping across all seven datasets (Fig. \ref{fig:kendall_tau}). A diagonal pattern --- where each ground-truth covariate is strongly and exclusively correlated only with its corresponding latent feature --- is robustly maintained for age (top row), gender (middle row), and site (bottom row) across all imaging modalities. These results provide a compelling evidence that MedIDL effectively performs feature disentanglement. 

\subsection{Comprehensive ablation and sensitivity analysis}
\label{sec:ablation}
\begin{figure*}[!t]
    \centering
    \includegraphics[width=\textwidth]{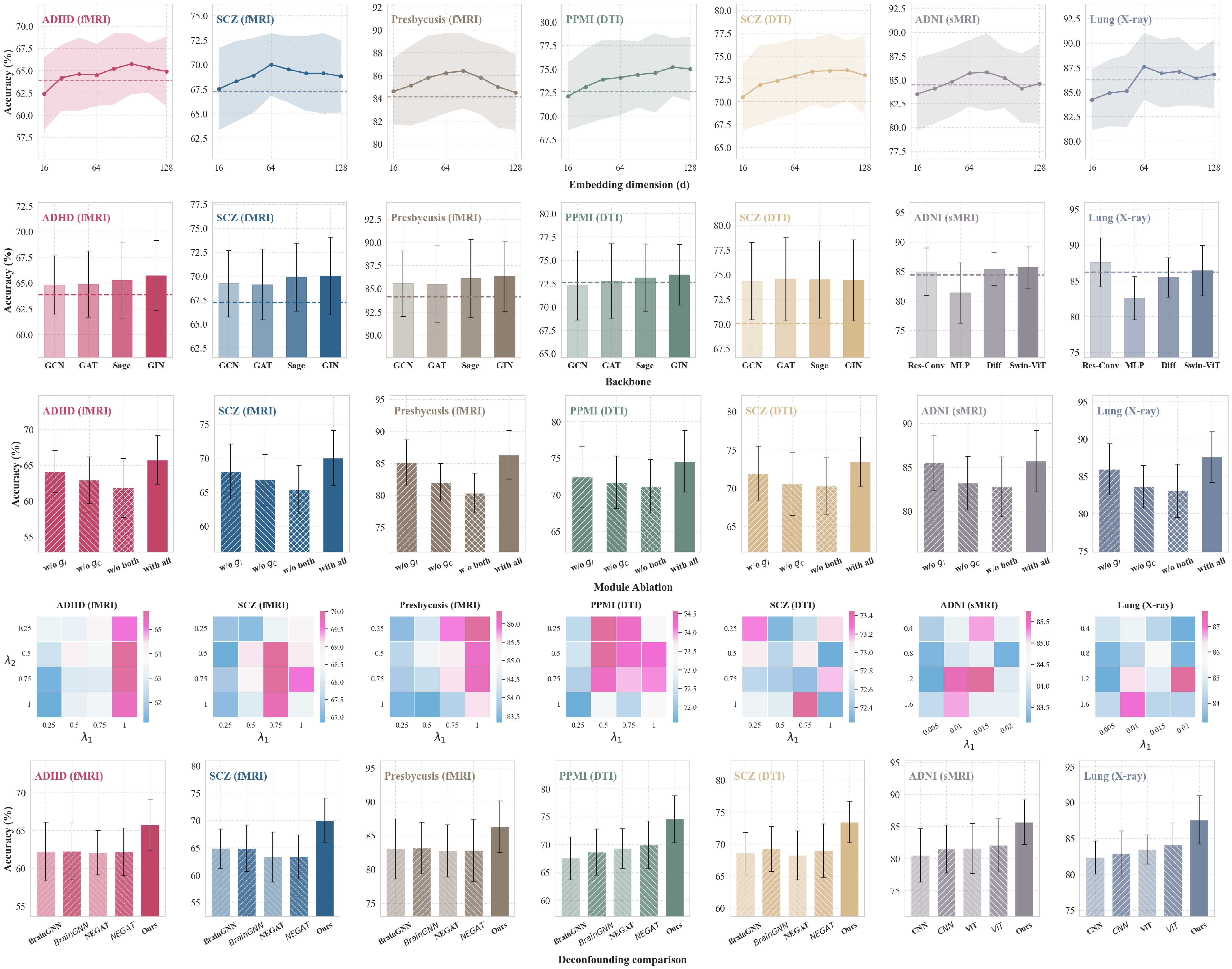}
    \caption{Comprehensive ablation and sensitivity analysis of the MedIDL framework across diverse imaging modalities. 
\textit{(Row 1)} Impact of the unified latent embedding dimension ($d$) on classification accuracy. Shaded regions indicate standard deviations. The dashed line denotes the top competing method.
    \textit{(Row 2)} Robustness validation across various modality-aware backbones. The dashed line denotes the top competing method.
    \textit{(Row 3)} Component-wise ablation of the tri-branch disentanglement module. 
    \textit{(Row 4)} Joint hyperparameter sensitivity heatmaps for the covariate alignment weight ($\lambda_1$) and latent reconstruction weight ($\lambda_2$), suggesting a stable, broad high-accuracy zone (darker red regions). 
    \textit{(Row 5)} Superiority of our implicit latent disentanglement (solid bars) over traditional input-level de-confounding via linear regression. Dark hatched bars represent explicit de-confounding (regressing out covariates prior to classification), while light hatched bars denote classification without covariate removal.
}
    \label{fig:ablation}
\end{figure*}

To validate the architectural components, optimization stability, and generalizability, we conducted extensive ablation and sensitivity studies (Fig.~\ref{fig:ablation}). 

\subsubsection{Information bottleneck and latent dimensionality}
The first row of Fig.~\ref{fig:ablation} delineates the sensitivity of the model with respect to the latent embedding dimension $d$. Across all datasets, a clear upward trajectory in accuracy is observed as the dimension increases from $16$. However, the performance typically reaches an optimal plateau between $d=32$ and $64$. Beyond $64$ (e.g., at $d=128$), the accuracy either saturates or marginally declines. This phenomenon aligns with the information bottleneck principle: overly expansive latent spaces encourage the memorization of redundant noise and increase the risk of overfitting. 

\subsubsection{Robustness across modality-aware backbones}
The second row of Fig.~\ref{fig:ablation} benchmarks the disentanglement performance across different backbone encoders. For functional and structural brain graphs, we evaluated Graph Convolutional Networks (GCN), Graph Attention Networks (GAT), GraphSAGE (Sage), and GIN. While GIN consistently yields slight advantages, the framework maintains highly stable performance across all graph encoders. Similarly, for 3D/2D image data, we substituted the backbone with residual convolutional networks (Res-Conv), basic MLPs, diffusion-based bottlenecks (Diff), and Swin-ViT. The absence of catastrophic performance drops across these diverse architectures confirms that our framework successfully purifies the embeddings independent of the upstream feature extractors.

\subsubsection{Effectiveness of the tri-branch disentanglement}
The third row of Fig.~\ref{fig:ablation} investigates the core components of our framework: the tri-branch disentanglement mechanism. We compared the full MedIDL model (``with all'') against three degenerative variants: removing the individual head (``w/o $g_I$''), removing the covariate head (``w/o $g_C$''), and removing both heads simultaneously (``w/o both'', reducing the framework to a standard encoder-decoder). Across all datasets, the ``w/o both'' variant suffers from the most severe performance degradation. In addition, removing the covariate head (``w/o $g_C$'') causes a sharper decline in accuracy than removing the individual head (``w/o $g_I$''). This empirically corroborates our hypothesis that demographic variables and site-specific biases act as the dominant confounders in medical imaging. 

\subsubsection{Hyperparameter stability}
The fourth row presents the accuracy heatmaps by varying the loss weighting hyperparameters ($\lambda_1$ for covariate alignment and $\lambda_2$ for reconstruction). 
The heatmaps reveal a consistent, broad ``sweet spot'' across the modalities. While the framework demonstrates remarkable tolerance to hyperparameter shifts, moderate-to-high regularization weights typically yield the best trade-off. 

\subsubsection{Superiority over explicit de-confounding strategies}
Finally, the fifth row of Fig.~\ref{fig:ablation} addresses a critical question: \textit{Is our implicit representation disentanglement superior to traditional linear regression-based de-confounding?} We compared conventional linear regression-based de-confounding against our framework. The empirical results are definitive. For both graph baselines and 3D/2D image baselines, standard linear regression fails to sufficiently mitigate confounding bias, yielding performance that is either marginally superior or virtually identical to uncorrected baselines. In contrast, our framework achieves significantly higher accuracies across all datasets. 


\begin{figure*}[!t]
    \centering
    \includegraphics[width=\textwidth]{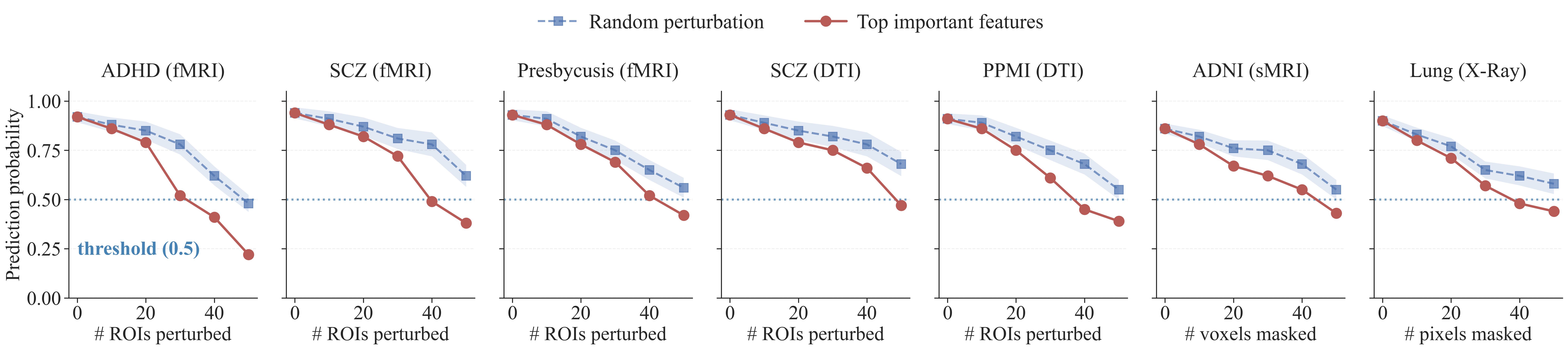}
 \caption{Quantitative validation of biomarker importance via counterfactual perturbation analysis across seven datasets. A plot depicts the degradation trajectories of the model's mean prediction probability as input features are progressively perturbed/masked (replaced by means from HCs) from the patients in the test sets. Solid red lines represent the targeted perturbation of the ``top important features'' identified by our framework. Dashed blue lines with shades serve as the baseline representing the mean and standard deviation from 5 ``random perturbations''. To accommodate the distinct spatial topologies of different modalities, the $x$-axis denotes the number of ROIs perturbed for fMRI and DTI, 3D voxels masked for sMRI, or 2D pixels masked for X-ray. 
 }
    \label{fig:counterfactual_perturbation}
\end{figure*}

\begin{figure*}[!t]
    \centering
    \includegraphics[width=0.95\textwidth]{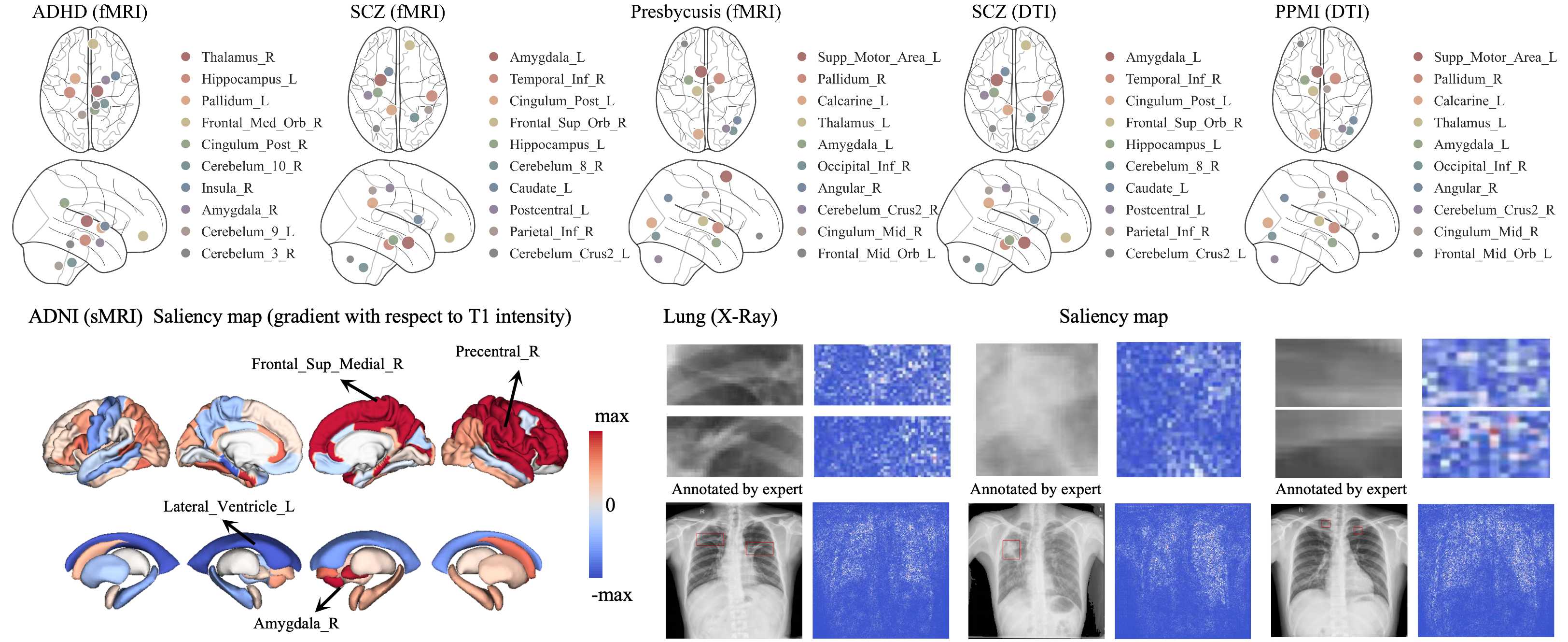}
\caption{Saliency maps identified by the proposed framework across seven datasets. \textit{(Top):} For graph-based brain networks, the framework maps the most discriminative ROIs in the MNI space, where the marker size scales proportionally with the node's importance score. \textit{(Bottom):} For voxel/pixel-based datasets, gradient-based feature attribution maps localize structural and radiological anomalies. On ADNI (sMRI), the gradient is with respect to T1 intensity after affine registration. Hence, positive values in the cortical region indicate gray matter atrophy; negative values in the ventricle indicate enlarged ventricle. On Lung (X-ray), red rectangles are pulmonary lesions annotated by a clinical expert. 
}
    \label{fig:explanation}
\end{figure*}
\subsection{Clinical interpretability and biomarker discovery}


To verify that our model extracts genuine pathological signals, we identify the disease-discriminating features. Specifically, we construct gradient-based saliency maps by computing the first-order partial derivatives of the predicted disease logits with respect to the input space (node features for graphs and pixel/voxel intensities for 2D/3D images). A counterfactual perturbation analysis by masking the identified top ROIs or voxels/pixels confirms that removing these signals leads to a more rapid decrease of the prediction probability on the patients from the test sets (Fig. \ref{fig:counterfactual_perturbation}). We further visualize these top ROIs/voxels/pixels in Fig. \ref{fig:explanation}. 

In the SCZ and ADHD cohorts, alterations are concentrated on the hippocampus, amygdala, and prefrontal regions, consistent with existing psychiatric literature~\cite{kim2025prefrontal}. On the PPMI (DTI) dataset, the identified regions strongly implicate basal ganglia and thalamic networks, in line with Parkinsonism degeneration~\cite{d2021thalamic}. On the ADNI (sMRI) dataset, the model highlights regions such as amygdala, lateral ventricle, medial superior frontal gyrus, precentral gyrus, which are well-established regions of early cognitive impairment~\cite{eskildsen2013prediction}. On the Lung (X-ray) dataset, the saliency maps delineate localized pulmonary opacities and lesions corresponding to Tuberculosis infections annotated by a clinical expert. 


\section{Discussion and Conclusion}
\label{sec:discussion}

In this study, we proposed a MedIDL framework to overcome a pervasive bottleneck in medical image analysis: the entanglement of disease-specific pathological signals with confounding covariates and individual variability. By introducing a modality-aware encoder alongside a tri-branch latent disentanglement architecture, MedIDL explicitly routes diagnostic features, covariate features, and individual variations into different latent spaces. Through extensive evaluations across seven datasets, our framework consistently demonstrated superior classification accuracy. Ablation, correlation, and counterfactual analyses confirmed that decoupling these representations effectively isolates demographic biases and improves classification accuracy and biomarker discovery.

Our MedIDL framework pushes the accuracy of fMRI-based classification for ADHD from 62--64\% \cite{zhang2025graph,chen2024self,shen2025disentangle} to 65.24\%. For SCZ and PPMI, our method outperformed the accuracies of upper 60\% to low 70\% from recent advanced GNNs~\cite{braingnn,negat,joao,graphcl} by achieving 70.02\% on SCZ (fMRI), 73.42\% on SCZ (DTI), and 74.46\% on PPMI (DTI). In the Presbycusis task, MedIDL achieved an impressive OvR-AUC of 87.05\%. On the ADNI (sMRI) dataset, we advanced the accuracy of classifying mild cognitive impairment (MCI) from HCs from 82--84\% \cite{yang2025diffmic,zhu2024vision} to 85.71\% and AUC to 86.48\%. On the Lung 2D X-ray dataset, our framework achieved an accuracy of 87.60\%, outperforming ViT and SimCLR by 4.12\% and 3.11\% respectively. These findings demonstrate that the performance bottleneck in contemporary medical image classification does not solely depend on the backbone architecture, but also on feature disentanglement. 


The identified disease-related regions align well with established clinical literature. On both the SCZ and ADHD datasets, the model identified hyper-connectivity and functional disruptions primarily within the default mode network, the prefrontal cortex, and the amygdala --- hubs intricately linked to cognitive control and emotional dysregulation in psychiatric literature~\cite{mcteague2017identification}. On the PPMI dataset, the top discriminative structural features were mapped to the basal ganglia, and thalamocortical tracts, which are the exact pathways compromised in dopaminergic depletion in Parkinson's disease~\cite{albin1989functional,schindlbeck2018network}. In the HC vs. MCI task, our model localized morphometric alterations to the right amygdala, neocortical hubs including the right superior medial frontal and precentral gyri, and the left lateral ventricle, capturing early limbic neurodegeneration accompanied by lateral ventricular enlargement, characteristic of prodromal Alzheimer's disease~\cite{jack2004comparison,johnson2012brain}. On the Lung (X-ray) dataset, our saliency maps delineated localized apical pulmonary opacities and cavitations, matching the radiological presentation of pulmonary tuberculosis~\cite{lyon2017pulmonary}. 

Despite its strong performance and generalizability, this study also has limitations. First, the covariate head relies on the availability of covariate information. If a covariate is missing, the framework lacks a mechanism to discover and decouple it. Second, we model the individual variability using a simple channel-wise Gaussian normalization. While this successfully absorbs random stochasticity, individual variability might not perfectly conform to a standard normal distribution. Finally, maintaining three parallel disentanglement branches and computing cross-subject similarity matrices increases the computational overhead and memory footprint during the training phase, although inference time remains unaffected.

In conclusion, we introduced the MedIDL framework, a highly adaptable, unified approach for extracting pure, disease-specific representations from complex medical imaging data. By adapting to diverse modalities and decoupling pathological signals from demographic covariates and individual variability, MedIDL establishes a new state-of-the-art across seven datasets. Beyond predictive accuracy, the framework isolated covariate features that only correlated with the designated true covariates and discovered biomarkers that aligned with established literature. Hence, MedIDL represents a significant step toward developing robust and interpretable artificial intelligence systems for real-world clinical deployment.

\section*{Acknowledgment}
This work was supported by grants from the National Natural Science Foundation of China (82441016, 82471940),  Natural Science Foundation of Shanghai (24TS1415000), and China Postdoctoral Science Foundation (2024M762010).

Data collection and sharing for this project were funded by the Alzheimer's Disease Neuroimaging Initiative (ADNI) (NIH Grant U01 AG024904; DOD ADNI W81XWH-12-2-0012). ADNI is funded by the National Institute on Aging, the National Institute of Biomedical Imaging and Bioengineering, and through contributions from numerous industry partners. A complete listing of ADNI investigators and funding sources can be found at: \url{http://adni.loni.usc.edu}.

Data used in the preparation of this article was obtained on July 21, 2025 from the PPMI database (\url{www.ppmi-info.org/access-dataspecimens/download-data}), RRID:SCR\_006431. For up-to-date information on the study, visit \url{www.ppmi-info.org}. PPMI -- a public-private partnership -- is funded by the Michael J. Fox Foundation for Parkinson's Research, and funding partners; including AbbVie, Aligning Science Across Parkinson's (ASAP), Avid Radiopharmaceuticals, Biogen, BioHaven, BioLegend, Bristol Myers Squibb, Celgene, Eli Lilly and Company, Genentech, GlaxoSmithKline, Golub Capital, Handl Therapeutics, Insitro, Janssen Neuroscience, Lundbeck, Merck \& Co., Inc., Meso Scale Discovery, Neurocrine Biosciences, Pfizer, Piramal, Roche, Sanofi Genzyme, Servier, Takeda, Teva, UCB, Verily, and Voyager Therapeutics.

We thank Dr. Wenting Rui at Huashan Hospital, Fudan University for ROI annotations on the Lung dataset.

\section*{Appendix}

An appendix is available in the online version.

\bibliographystyle{IEEEtran}
\bibliography{refs}

\end{document}


\bstctlcite{TMIreferencecontrol}
\maketitle
\renewcommand{\thefootnote}{\arabic{footnote}}

\section*{Appendix A: Architectural details}

\subsection*{GNN for brain graphs}

For brain graphs $\{ x_i = (A_i,X_i):i=1,\dots,B\}$ in a batch, the default encoder is a 2-layer \textit{Graph Isomorphism Network (GIN)} with the update of each layer being
\begin{equation}
\begin{gathered}
H_i^{(l)} = \operatorname{MLP}^{(l)}\Bigl(((1 + \epsilon)I + A_i )H_i^{(l-1) }\Bigr), \quad 
l = 1,2,
\end{gathered}
\end{equation}
where $H_i^{(l)}=[\mathbf{h}_{i,1}^{(l)},\dots,\mathbf{h}_{i,N}^{(l)}]^{\top}$ is the node features at the $l$-th layer of the $i$-th sample ($H_i^{(0)} = X_i$), $I$ is the identity matrix, \(\epsilon = 0\), and the multi-layer perceptron (MLP) at layer \(l\) transforms each node feature by
\begin{equation}
\begin{aligned}
&\operatorname{MLP}^{(l)}(\mathbf{h}_{i,v}^{(l)}) 
 = W_2^{(l)} \operatorname{ReLU}\Bigl(W_1^{(l)} \mathbf{h}_{i,v}^{(l)} + b_1^{(l)}\Bigr) + b_2^{(l)},
\end{aligned}
\end{equation}
with \(W_1^{(l)} \in \mathbb{R}^{d_{\text{hidden}} \times d}\), \(W_2^{(l)} \in \mathbb{R}^{d \times d_{\text{hidden}}}\), \(b_1^{(l)}, b_2^{(l)}\) being learnable parameters, ReLU is the activation function, and \(d_{\text{hidden}} = 64\). After the second layer, global average pooling yields
\begin{equation}
z_i = \frac{1}{N} \sum_{v=1}^{N} \mathbf{h}_{i,v}^{(2)}.
\end{equation}
For ablation studies we also support Graph Attention Net (GAT), GraphSAGE, and Graph Convolutional Network (GCN) with identical layer counts and hidden dimensions.

\subsection*{3D ViT for volumetric MRI}

To process 3D MRI volumes, we convert each 3D image $x_i$ into patches $\{x_{i,k}^{\text{patch}}\}_{k=1}^P$ and adopt a 3D ViT architecture with a patch size of $16 \times 16 \times 16$, where $P$ is the total number of 3D spatial patches. The 3D patch embedding for the $k$-th patch is formulated as
\begin{equation}
\begin{gathered}
p_{i,k} = \operatorname{Linear}\left(x^{\text{patch}}_{i,k}\right) + E^{\text{pos}}_{i,k}, \quad 
k = 1,\dots,P,
\end{gathered}
\end{equation}
where $E^{\text{pos}} \in \mathbb{R}^{P \times D}$ denotes the learnable positional embeddings. 

To feed these patch embeddings into the Transformer encoder, we construct an initial token sequence matrix $Z_i^{(0)} = [p_{i,1}, p_{i,2}, \dots, p_{i,P}]^\top \in \mathbb{R}^{P \times D}$. The backbone consists of an 8-layer Transformer encoder. For each layer $l = 1, \dots, 8$, the feature updates via MLPs, multi-head self-attention, layer normalization, and residual connections are formulated as:
\begin{equation}
\begin{aligned}
&Z_i^{(l-\frac{1}{2})} = Z_i^{(l-1)}+ \operatorname{MHSA}\left(\operatorname{LayerNorm}(Z_i^{(l-1)})\right),
\end{aligned}
\end{equation}
\begin{equation}
\begin{aligned}
Z_i^{(l)} &= Z_i^{(l-\frac{1}{2})} 
 + \operatorname{MLP}\left(\operatorname{LayerNorm}(Z_i^{(l-\frac{1}{2})})\right).
\end{aligned}
\end{equation}
Specifically, the encoder employs 8 Transformer layers with 6 attention heads per layer and a hidden dimension of $D = 768$. The global representation $z_{\text{g}}$ --- obtained via global average pooling over the final output patch tokens $Z_i^{(8)}$ --- is subsequently projected by a linear layer $W_{\text{proj}} \in \mathbb{R}^{64 \times 768}$ to derive the final representation vector:
\begin{equation}
    z_i = W_{\text{proj}} \cdot z_{\text{g}}.
\end{equation}

\subsection*{2D CNN for X-ray}

We adapt a residual neural network backbone to extract structural feature representations from 2D X-ray radiographs. To accommodate continuous spatial down-sampling and channel expansion across residual stages, the residual block is formulated as:
\begin{equation}
\begin{aligned}
&\mathbf{h}_i^{(l+1)} = \mathcal{W}\left(\mathbf{h}_i^{(l)}\right) 
+ \operatorname{Conv2D}_1\Big(\operatorname{ReLU}\big(\operatorname{BN}\big(\operatorname{Conv2D}_2(\mathbf{h}_i^{(l)})\big)\big)\Big),
\end{aligned}
\end{equation}
where $\mathbf{h}_i^{(l)}$ denotes the input feature map at layer $l$ ($\mathbf{h}_i^{(0)}=x_i$), and $\operatorname{BN}$ represents batch normalization. $\mathcal{W}(\cdot)$ is the projection shortcut --- implemented as a $1 \times 1$ convolution with a stride of $2$. $\operatorname{Conv2D}_j$ performs spatial 2D convolution with a $3 \times 3$ kernel and a stride of $j$. After four down-sampling stages, a global average pooling layer aggregates the feature maps into a $d$-dimensional embedding $z_i$.

\section*{Appendix B: Dataset details and preprocessing}

\label{sec:results}

Details of the 7 datasets are in Table~\ref{tab:data_info}. Their preprocessing steps are given below.

\begin{table*}[!t]
\centering
\caption{Demographic, clinical, and detailed imaging acquisition characteristics across the seven datasets. }
\label{tab:data_info}
\renewcommand{\arraystretch}{1.3}
\small
\setlength{\tabcolsep}{4pt}

\begin{tabularx}{\textwidth}{
>{\raggedright\arraybackslash}m{2.1cm}
>{\raggedright\arraybackslash}m{1.7cm}
>{\centering\arraybackslash}m{0.8cm}
>{\centering\arraybackslash}m{1.1cm}
>{\centering\arraybackslash}m{1.3cm}
>{\centering\arraybackslash}m{1.7cm}
>{\raggedright\arraybackslash}m{2.2cm}
X
}
\toprule
{\makecell[l]{Dataset \&\\Modality}} & 
{Class} & 
{Sites} & 
{Subjects} & 
{\makecell[c]{Gender\\(F/M)}} & 
{\makecell[c]{Age\\($\text{mean}\pm\text{std}$)}} &
{\makecell[c]{Scanner\\Hardware}} &
{\makecell[c]{Acquisition\\Parameters}} \\
\midrule

ADHD-200 (\textit{fMRI}) & \makecell[l]{ADHD\\HC} & 4 & \makecell[c]{275\\205} & \makecell[c]{79/196\\105/100} & \makecell[c]{11.5$\pm$2.8\\12.1$\pm$3.2} & Siemens/Philips 3T & TR = 2000/2500 ms, TE = 15--30 ms, FA = 75$^\circ$/90$^\circ$, voxel size $\approx$ 3 mm$^3$ \\
\hline

SCZ (\textit{fMRI}) & \makecell[l]{SCZ\\HC} & 3 & \makecell[c]{137\\190} & \makecell[c]{62/75\\92/98} & \makecell[c]{24.8$\pm$7.5\\24.9$\pm$8.2} & 3.0T Siemens Magnetom Trio Tim scanner & TR = 2 s, TE = 30 ms, flip angle = 90$^\circ$, FOV = 220 $\times$ 220 mm$^2$ \\
\hline

Presbycusis (\textit{fMRI}) & \makecell[l]{Presbycusis\\N-PTA\\HC} & 1 & \makecell[c]{130\\154\\112} & \makecell[c]{55/75\\70/84\\48/64} & \makecell[c]{63.2$\pm$3.3 \\63.6$\pm$4.6\\63.1$\pm$3.2} & 3.0T Philips Achieva MR scanner & TR = 2 s, TE = 35 ms, FOV = 240 $\times$ 240 mm$^2$, resolution = 3.75 $\times$ 3.75 mm$^2$, 35 slices, thickness = 4 mm, 240 vols \\
\hline

PPMI (\textit{DTI}) & \makecell[l]{PD\\HC} & 15 & \makecell[c]{152\\41} & \makecell[c]{103/49\\28/13} & \makecell[c]{61.3$\pm$9.1\\59.5$\pm$10.8} & 3.0T Siemens Tim Trio scanner & TR/TE = 900/88 ms, FA = 90$^\circ$, 72 slices, $b$ = 1000 s/mm$^2$, 64 diffusion dirs, voxel size = 2 $\times$ 2 $\times$ 2 mm$^3$, Cardiac-triggered \\
\hline

SCZ (\textit{DTI})\textsuperscript{$\dag$} & \makecell[l]{SCZ\\HC} & 3 & \makecell[c]{137\\190} & \makecell[c]{62/75\\92/98} & \makecell[c]{24.8$\pm$7.5\\24.9$\pm$8.2} & 3.0T Siemens Magnetom Trio Tim scanner & \makecell[l]{$b$ = 1000 s/mm$^2$, 32 dirs,\\ resolution = 2$\times$2$\times$2 mm$^3$} \\
\hline

ADNI (\textit{sMRI}) & \makecell[l]{MCI\\HC} & 4 & \makecell[c]{201\\238} & \makecell[c]{101/100\\125/113} & \makecell[c]{76.5$\pm$7.0\\73.7$\pm$8.0} & Multi-site 1.5T/3T & \makecell[l]{T1-weighted MPRAGE,\\ isotropic 1 mm$^3$ voxels} \\
\hline

Lung (\textit{2D  X-Ray}) & \makecell[l]{TB\\HC} & 1 & \makecell[c]{204\\300} & \makecell[c]{72/132\\98/202} & \makecell[c]{33.7$\pm$14.9\\33.4$\pm$14.0} & Philips DR Digital Diagnost system & Frontal projection (PA/AP), approx. 3K $\times$ 3K pixels, PNG format \\

\bottomrule
\end{tabularx}

\begin{flushleft}
\small
\textsuperscript{$\dag$} \textit{The SCZ (DTI) dataset has the same sample size and characteristics as the SCZ (fMRI) dataset. ADHD: attention deficit hyperactivity disorder. SCZ: schizophrenia. N-PTA: normal pure-tone audiometry. PD: Parkinson's disease. MCI: mild cognitive impairment. TB: tuberculosis. HC: healthy control.}
\end{flushleft}
\end{table*}


\subsection*{Brain graph preprocessing (fMRI and DTI)}


For functional connectivity, rs-fMRI data are parcellated using the AAL1 atlas~\cite{aal}, resulting in $N=116$ ROIs. Each ROI is treated as a node, while the pairwise functional connectivity between ROIs defines the edges, forming the adjacency matrix. Node features are characterized by three ALFF measures --- Slow-5 (0.01--0.027 Hz), Slow-4 (0.027--0.073 Hz), and the conventional band (0.01--0.08 Hz) --- which quantify intrinsic neuronal oscillations through the Fourier transform of BOLD time series~\cite{frequency1}. Consequently, each node is represented by a 3-dimensional ALFF feature vector.

For structural connectivity, DTI data are aligned to the same AAL1 template to ensure ROI correspondence with the rs-fMRI data. The mean streamline number between two ROIs is employed as the edge weight, reflecting the strength of white matter connectivity. To characterize localized tract microstructures, four core diffusion tensor metrics --- FA, MD, axial diffusivity, and RD --- are computed and concatenated alongside tensor MO and TR, yielding a 6-dimensional feature vector for each node.

To ensure stable learning across modalities, node features are normalized to $[0, 1]$ and edge weights to $[-1, 1]$.

\subsection*{3D volumetric MRI preprocessing}
For 3D sMRI (T1-weighted scans from ADNI), the raw neuroimaging volumes underwent a rigorous standardization pipeline utilizing the FMRIB Software Library (FSL) package \cite{jenkinson2012fsl}. The preprocessing protocol incorporated automated skull-stripping to isolate brain tissues, followed by affine registration ---FMRIB's Linear Image Registration Tool (FLIRT) --- to the standard MNI152 stereotaxic space to eliminate anatomical misalignment. Finally, intensity z-score normalization was applied and the volumes were center-cropped and down-sampled to a uniform grid of $96 \times 96 \times 96$ voxels. 
%

\subsection*{2D planar X-ray preprocessing}
For 2D planar images (chest radiographs), the raw X-ray scans inherently suffer from varied spatial resolutions and exposure discrepancies. To standardize the inputs, all images were strictly resized to a fixed spatial dimension of $224 \times 224$ pixels using bicubic interpolation. Subsequently, global min-max normalization was applied to map the pixel intensities into the $[0, 1]$ range. This normalization step homogenizes the contrast distribution, ensuring numerical stability for the downstream encoders.

\section*{Appendix C: Evaluation strategy}

To demonstrate the efficacy of our framework, comprehensive benchmarking was conducted against multiple state-of-the-art baselines across graph and Euclidean modalities. For fMRI and DTI graph classification, we evaluated our model against nine competing methods, which are categorized into: (i) two supervised learning methods, namely BrainGNN \cite{li2021braingnn} and NEGAT \cite{negat}; (ii) six self-supervised learning (SSL) methods, including GraphCL \cite{graphcl}, JOAO \cite{joao}, LaGraph \cite{lagraph}, AGCL \cite{agcl}, GATE \cite{gate}, and BrainGCL \cite{braingcl}; and (iii) one graph disentangling framework, DMG \cite{dmg}. For volumetric and planar images, our model was compared against standard deep vision backbones. Specifically, for 3D medical image classification, the baselines comprised vanilla CNN~\cite{li2014medical}, ResNet~\cite{he2016deep}, ViT~\cite{liu2021swin}, GF-Net~\cite{zhang20223d}, V-Mamba~\cite{visionmambda}, DiffMed-v2~\cite{yang2025diffmic}, SimCLR~\cite{simclr}, ADIOS~\cite{shi2022adversarial}, and SGLA-Net~\cite{jiang2025segmentation}. For 2D images, the framework was benchmarked against the 2D counterparts of the aforementioned 3D backbones.

For a fair comparison, all baselines take the officially released code or are re-implemented. All methods are evaluated with identical training/validation/test splits (60/20/20\%) in 5-fold cross-validation. In supervised learning, the validation set is used to determine the optimal training epoch for evaluating the test set. In SSL, the validation set serves to tune the hyperparameter of the downstream classifier --- a linear support vector machine (SVM). Evaluation metrics are accuracy and area under the curve (AUC) (OvR-AUC for multi-class in Presbycusis).

\bibliographystyle{IEEEtran}
\bibliography{refs}